\documentclass[journal]{IEEEtran}

\usepackage{amsmath,amsfonts}
\usepackage{algorithmic}
\usepackage{algorithm}
\usepackage{array}
\usepackage[caption=false,font=normalsize,labelfont=sf,textfont=sf]{subfig}
\usepackage{textcomp}
\usepackage{stfloats}
\usepackage{url}
\usepackage{verbatim}

\usepackage{cite}
\usepackage{booktabs} 
\usepackage{array}    
\usepackage{multirow}
\usepackage[dvipdfmx]{graphicx}
\usepackage{amsmath}
\usepackage{amssymb}  
\usepackage{hyperref}
\hypersetup{hidelinks,
	colorlinks=true,
	allcolors=black,
	pdfstartview=Fit,
	breaklinks=true}

\ifCLASSINFOpdf
\else
\fi
\begin{document}
%
%
%
%
\title{Rule-Based Conflict-Free Decision Framework \\ in Swarm Confrontation}

%

\author{Zhaoqi Dong, Zhinan Wang, Quanqi Zheng, Bin Xu, Lei Chen, ~\IEEEmembership{Member,~IEEE}
and Jinhu L\"u, ~\IEEEmembership{Fellow,~IEEE}

\thanks{
This work was supported by the National Science Foundation of China under Grants 62088101, 62003015
\textit{(Corresponding author: Lei Chen)}.
}
\thanks{Zhaoqi Dong, Quanqi Zheng and Lei Chen are with the Advanced Research Institute of Multidisciplinary Sciences and State Key Laboratory of CNS/ATM, Beijing Institute of Technology, Beijing 100081, China (e-mail: dongzhaoqi@bit.edu.cn; zqq\_bit@bit.edu.cn; bit\_chen@bit.edu.cn).}
\thanks{Bin Xu and Zhinan Wang are with the School of Machanical Engineering, Beijing Institute of Technology, Beijing 100081, China (e-mail: bitxubin@bit.edu.cn; wangzn@bit.edu.cn).}
\thanks{Jinhu L\"u is with the School of Automation Science and Electrical Engineering, Beihang University, Beijing 100191, China (e-mail: jhlu@iss.ac.cn).}
\thanks{The experiment video is available at \url{https://b23.tv/ozlwUAs}.}
\thanks{The code is available at \url{https://github.com/dddddzq/Swarm-confrontation}.}
}

%
%

\markboth{Journal of \LaTeX\ Class Files,~Vol.~14, No.~8, August~2015}%
{Shell \MakeLowercase{\textit{et al.}}: Bare Demo of IEEEtran.cls for IEEE Journals}
%



\maketitle
\begin{abstract}
Traditional rule--based decision--making methods with interpretable advantage, such as finite state machine, suffer from the jitter or deadlock(JoD) problems in extremely dynamic scenarios. 
To realize agent swarm confrontation, decision conflicts causing many JoD problems are a key issue to be solved.
Here, we propose a novel decision--making framework that integrates probabilistic finite state machine, deep convolutional networks, and reinforcement learning to implement interpretable intelligence into agents. 
Our framework overcomes state machine instability and JoD problems, ensuring reliable and adaptable decisions in swarm confrontation. 
The proposed approach demonstrates effective performance via enhanced human--like cooperation and competitive strategies in the rigorous evaluation of real experiments, outperforming other methods.
\end{abstract}


%
\IEEEpeerreviewmaketitle

\section{Introduction}
\IEEEPARstart{A}{s} artificial intelligence advances, autonomous unmanned systems hold great potential for the future in both civil\cite{kaufmann2023champion,10530429} and military fields \cite{10472871}.
Due to the real--world requirements, swarms consisting of autonomous unmanned systems are an riveting topic in research fields\cite{9894112,zhou2022swarm,10423569}.
To realize their abilities, especially in swarm confrontation scenario\cite{haarnoja2024learning}, autonomous decision--making is a fundamental problem for the purpose of achieving victory.
Unlike strategy planning in a single agent, strategies among the agents need to formulate as the decisions must consider the cooperative actions of agents\cite{10534867,10517396} and the resolutions of conflicts\cite{10418997} caused by the planned strategies, not to mention that the opponents' strategies should be carefully predicted and concerned\cite{10433866}.
Hence the establishment of an effective decision--making framework in adversarial environment can promote the combat ability for swarms which should plan the strategies with more stability for cooperation, more robustness for conflictions, and more intelligence for confrontation\cite{nashed2022survey}.

\begin{figure} 
    \centering
    \includegraphics[width=0.5\textwidth,height=0.2185\textwidth]{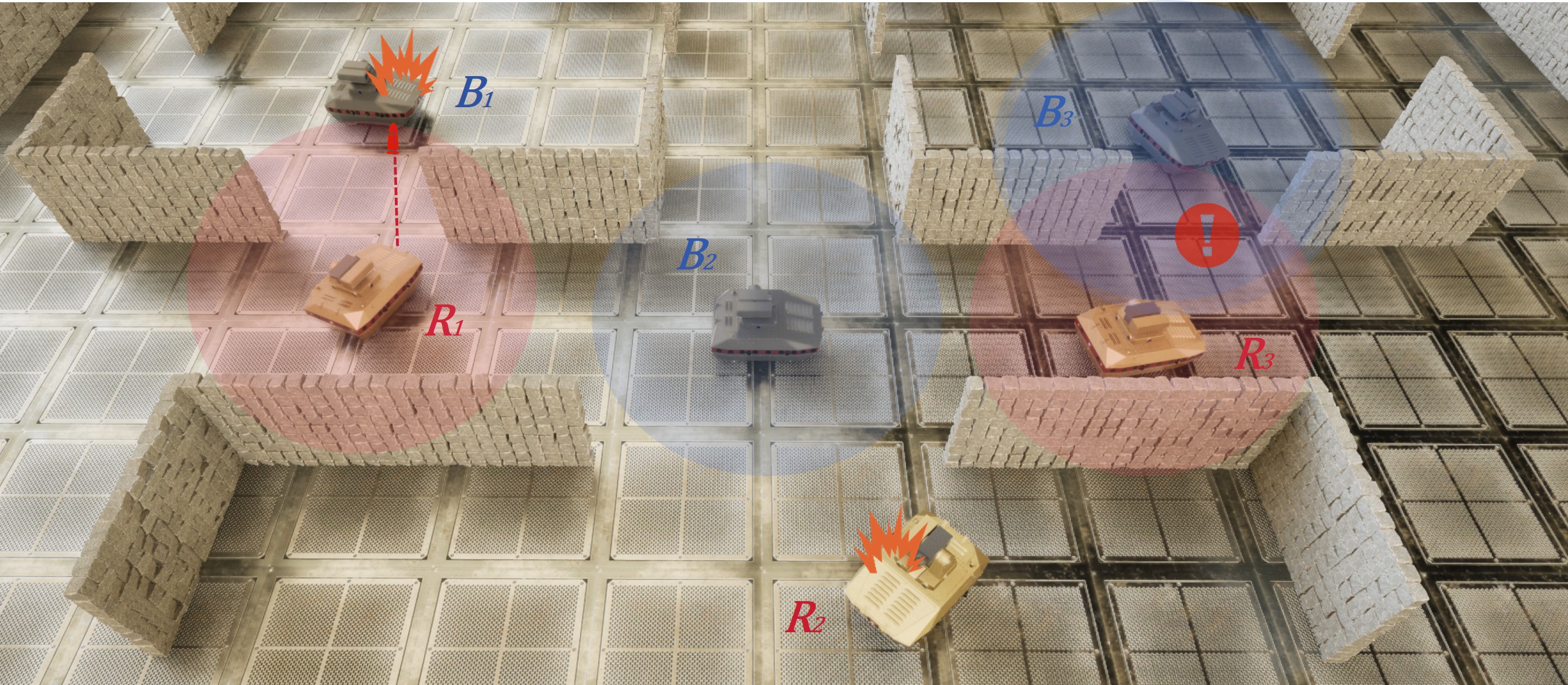}
    \caption{Two teams of agents, marked by red and blue colors are battling in a environment with obstacles. The colored shallow circles represent the detect range facilitated by the detectors. The agent reveals the other agents if they enter its detect circle, denoted with the exclamation point. Once an agent is detected by other agents, it may be destroyed by the missiles of the others, such as the agents marked with the fire symbols. 
    }
    \label{environment}
\end{figure}
Although machine learning methods\cite{10120732,paredes2024fully} demonstrate that they can handle the uncertainty in the confrontation scenario via online decision–making facilitated by deep learning networks and enormous collected data, the non-interpretable nature\cite{rudin2019stop} limits their applications in the real world for we do not expect some bizarre behaviors in the battlefield out of our common principles. 
Meanwhile, rule–based methods\cite{10149438,ma2021rule} provide another plausible solution for swarm confrontation. On one hand, their explainability makes sure every decision and control delicately reflect the command and action in human behaviors.
On the other hand, they show more reliability and robustness after development in decades\cite{aksjonov2023safety}, conquer the current artificial intelligence which struggles to implementations. 
Rule-based decision-making methods solve many problems with expert experience and tactical manuals. 
Traditionally, they employ the if–else–then logic\cite{9199531} to build a link from predefined conditions to corresponding actions, which work well in predefined tasks.
It is obvious that we can not use this logic to realize real applications nowadays for the environment faced by an agent is too complicated to be modeled.
Furthermore, it is not suitable for embedding artificial intelligence since a single action is perfectly designed to a single condition.
An alterative way is to build a link from condition states to action states, where several actions forming a sequence are banded into a state, giving birth to a new solution called by finite state machine\cite{10193849, 10149438, 8936943}. 
Given automatic driving fields crying out for online decision--making, \cite{10192480} proposes the finite state machine with linear time logic formulas to intelligent transportation for handling the demands of continuous behavior planning in dynamic environments. 
Meanwhile, for military uses, a framework\cite{10149438} combines the finite state machine and condition--action rule in order to reduce the model complexities caused by adversarial characteristics.
Sooner, they further promote it to aircraft battlefield\cite{10674773} in the form of a closed--loop state transition and achieve the observation--order--decision--action loop successfully completed.

With the widely use of finite state machine, a key issue, however, arises.
In the swarm confrontation scenario, the frequently decision conflicts result in state jitter and deadlock problems\cite{zakir2022robot}.
Unlike the state transitions for a single agent or a simple circumstance, where the triggering condition is fixed and easy to retrieve, these transitions occur rapidly in the swarm not only in an agent but also among agents.
It, hence, leads that the decision process must consider multiple conditions that often conflict with each other\cite{shen2024opinion,10412631}.
Consequently, an agent sometimes do not know what to do next, meaning the decision jittering among states or deadlocking one state\cite{9777253}.  
To overcome this obstacle, a simple solution is to introduce the time interval.
Some employ new transiting states to recuperate the defect status\cite{10260682} for the jitter and deadlock, allowing the agent rapidly slipping out from the dilemma even though they bring out model complexities from another perspective. 
Others induce `thinking time' for the agent in the dilemma\cite{9777253}. 
Given enough decision time, a new optimal solution can be somehow searched but it may not punctual enough to the circumstance.
Besides these stalling tactics, a more promising approach is to reconcile conflicts.
For example, \cite{9919233} fusions the conditions, especially for those conflicted ones, into a united state meaning that it rather chooses to understand the situation clearly first than makes a struggling decision first.
When the situation becomes too complex, deep learning\cite{10015270,9827367} is induced to approximate the situation curve that can not be modeled, assisting the agent recognize emergent circumstances.
This method facilitates the decision process via data--driven approach, although it also brings in non--interpretability to the observing side.

The task, however, becomes more tedious when it comes to the swarm confrontation scenario where the agent need to make decisions not only considering the environment and the opponents but also concerning the allies. 
The mentioned above methods can barely handle this complicated situation. 
Hence, we propose a decision--making framework that aims to solve the jitter and deadlock problems caused by condition conflicts.
The proposed framework can provide accurate and smooth decision--making ability for the agents in swarm confrontation.
Specifically, they are as follows:
\begin{enumerate} 
\item We translate the constant condition matrix into probabilistic ones via fusing the condition features and successfully eliminate the jitter and deadlock in the finite state machine.
\item To establish a transition probability matrix is not always easy for a complicated scenario, such as swarm confrontation.
We develop a deep learning method to approach the transition probability matrix instead of constructing its explicit form.
\item We employ the reinforcement learning method to optimal the transition probability matrix in order to promote the intelligence ability that may be declined by the absence of explicit transition matrix, resolve internal conflicts and guarantees the decision--making of the agent is accurate in time.
\end{enumerate}

We organize the rest of this paper as follows.
Section \ref{sec:problem_formulation} introduces the swarm adversarial problem and models the agents.
Section \ref{sec:rc-ppo_net}  presents the multilevel adversarial decision framework, providing details of the decision framework design.
Section \ref{sec:results_discussions} presents comparative experiments, simulation results, and real experiments.
Section \ref{sec:conclusion} concludes the paper.

\section{Problem Formulation and Preliminaries}
\label{sec:problem_formulation}

\begin{figure} 
    \centering
    \includegraphics[width=0.3\textwidth,height=0.2904\textwidth]{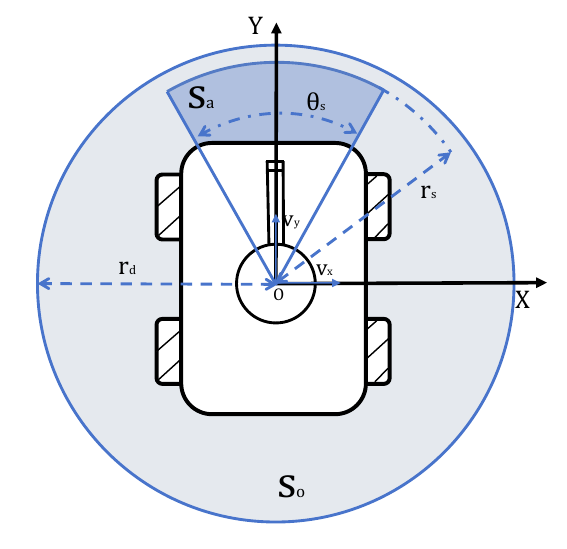}
    \caption{The model represents the agent and its sensor system within a two-dimensional coordinate framework.
    \( O \) marks the center of agent, which serves as the origin of the \( X \)-\( Y \) coordinate system. 
    The blue circular area \( S_o \) represents the perception range, while \( S_a \) denotes the attack range aligned with the velocity vector \( \boldsymbol{v} \).}
    \label{formulation}
\end{figure}

Swarm confrontation involves two independed teams, each consisting several agents equipped with detectors and missiles, and battles to win in an environment full of obstacles. 
We rule that a single agent is destroyed once it is detected and hit with a missile launched by the rivals. 
The confrontation ends with one team being totally destroyed.
Agents must make real--time decisions to adapt the dynamic and adversarial environment, as shown in Fig.~\ref{environment}.

In order to characterize the motion of each agent in this environment, we adopt a double--integrator system.
The dynamics of agent $i$ are given by
\begin{align}
\dot{\boldsymbol{p}}_i &= \boldsymbol{v}_i, \\
\dot{\boldsymbol{v}}_i &= \boldsymbol{u}_i,
\end{align}
where \( \boldsymbol{p}_i \in \mathcal{R}^2 \), \( \boldsymbol{v}_i \in \mathcal{R}^2 \), and \( \boldsymbol{u}_i \in \mathcal{R}^2 \) denote the position, velocity, and control input of agent \( i \), respectively. 
\( \mathcal{R}^2 \) represents the field of real numbers.
The perception and attack capabilities of each agents are defined by its perception range \( S_o \) and attack range \( S_a \) as:
\begin{equation}
S_o = \{(x, y) \in \mathcal{R}^2 \mid \| (x - x_i, y - y_i) \|_E \leq r_d \},
\end{equation}
\begin{equation}
\begin{split}
    S_a = \{(x, y) \in \mathcal{R}^2 \mid \| (x - x_i, y - y_i) \|_E \leq r_s, \\ 
    \angle(\boldsymbol{v}_i, (x - x_i, y - y_i)) \leq \theta_s / 2 \},
\end{split}
\end{equation}
where \( r_d > r_s \) ensures that agents can detect targets before engaging them, as shown in Fig.~\ref{formulation}. 
The notation \( \angle(\boldsymbol{v}_i, x)\) denotes the angle between the velocity \( \boldsymbol{v}_i \) and the relative position \( (x - x_i, y - y_i) \), measured in the range \([0, \pi]\).
The attack direction of the agent forms an angular sector aligned with \(\boldsymbol{v}_i\).
Interactions between agents are determined by the relative distance \( d_{ij} \) and relative angle \( \theta_{ij} \) as:
\begin{align}
d_{ij} &= \| \boldsymbol{p}_j - \boldsymbol{p}_i \|_E, \\
\theta_{ij} &= \arccos\left(\frac{(\boldsymbol{p}_j - \boldsymbol{p}_i)^\top\ \cdot \boldsymbol{v}_i}{\| \boldsymbol{p}_j - \boldsymbol{p}_i \|_E  \| \boldsymbol{v}_i \|_E}\right),
\end{align}
where \( d_{ij} \) represents the Euclidean distance between agents \( i \) and \( j \), and \( \theta_{ij} \) quantifies the angular alignment between \( \boldsymbol{v}_i \) and \( \boldsymbol{p}_j - \boldsymbol{p}_i \).

\begin{figure*}
    \centering   
    \includegraphics[width=1.0\textwidth,height=0.5020024809498\textwidth]{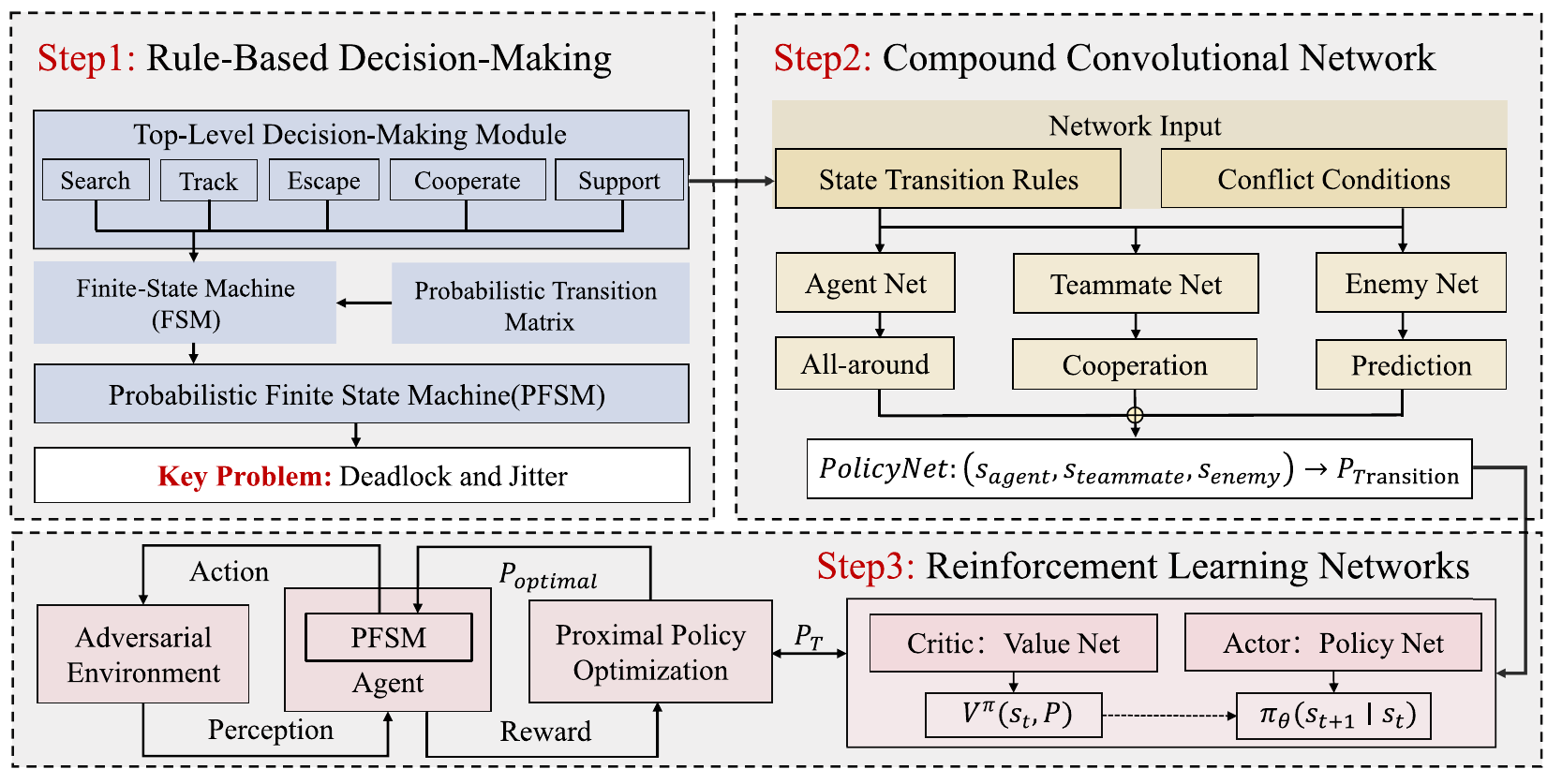}
    \caption{Decision--making of the single agent includes three steps: rule-driven decision-making, composite convolutional network, and reinforcement learning network. Each step is equipped with detailed subcomponents and descriptions, reflecting a structured intelligent decision--making system architecture.}
    \label{overview}
\end{figure*}

Although each agent follows the simple motion model described above, collective swarm confrontation introduces additional complexity arising from adversarial interactions and environmental constraints. These factors influence agent behavior in unpredictable ways. In particular, agents encounter rapidly changing conditions, such as the movement of opposing agents, real-time variations in the environment, and the ongoing need to coordinate with teammates. Consequently, agents make real--time decisions that balance multiple conflicting objectives in a highly dynamic and uncertain setting.
The primary challenge in this environment is the concurrency of decision triggers. Each agent must handle multiple tasks simultaneously, and these tasks often involve mutually competing conditions--satisfying one objective may impede the achievement of another. The coexistence of these conflicting demands leads to decision instability, manifested as jitter and deadlock. Jitter occurs when agents repeatedly switch between incompatible actions due to competing requirements, resulting in instability that prevents them from settling into a stable, effective strategy. Deadlock arises when agents become trapped in a particular state, unable to transition to more optimal or necessary actions because of unresolved conflicts among tasks. Both jitter and deadlock substantially diminish swarm efficiency and performance, as they prevent agents from executing coherent strategies under adversarial conditions.

The core reason for these issues is the lack of a mechanism that effectively resolves conflicting tasks within the decision-making process. Balancing multiple objectives under rapidly changing conditions demands a more sophisticated framework that ensures agents can maintain stable, effective decisions despite environmental uncertainty. Therefore, we develop a decision-making framework for swarm confrontation. This framework integrates various environmental factors, addresses competing decision triggers, and ensures that agents adapt to real-time changes without succumbing to instability.

\section{Methodology}
\label{sec:rc-ppo_net}

The proposed framework addresses decision--making challenges in swarm confrontation by integrating a probabilistic finite state machine (PFSM), composite convolutional networks, and reinforcement learning, as shown in Fig.~\ref{overview}. 
The PFSM governs state transitions through a probabilistic matrix \( \boldsymbol{P} \), dynamically generated by convolutional networks that unify observations from agent, teammate, and enemy streams. 
Reinforcement learning optimizes \( \boldsymbol{P} \) to mitigate jitter and deadlock.


\subsection{Rule--Based Decision--Making}
\begin{figure} 
    \includegraphics[width=0.5\textwidth,height=0.3602709986524\textwidth]{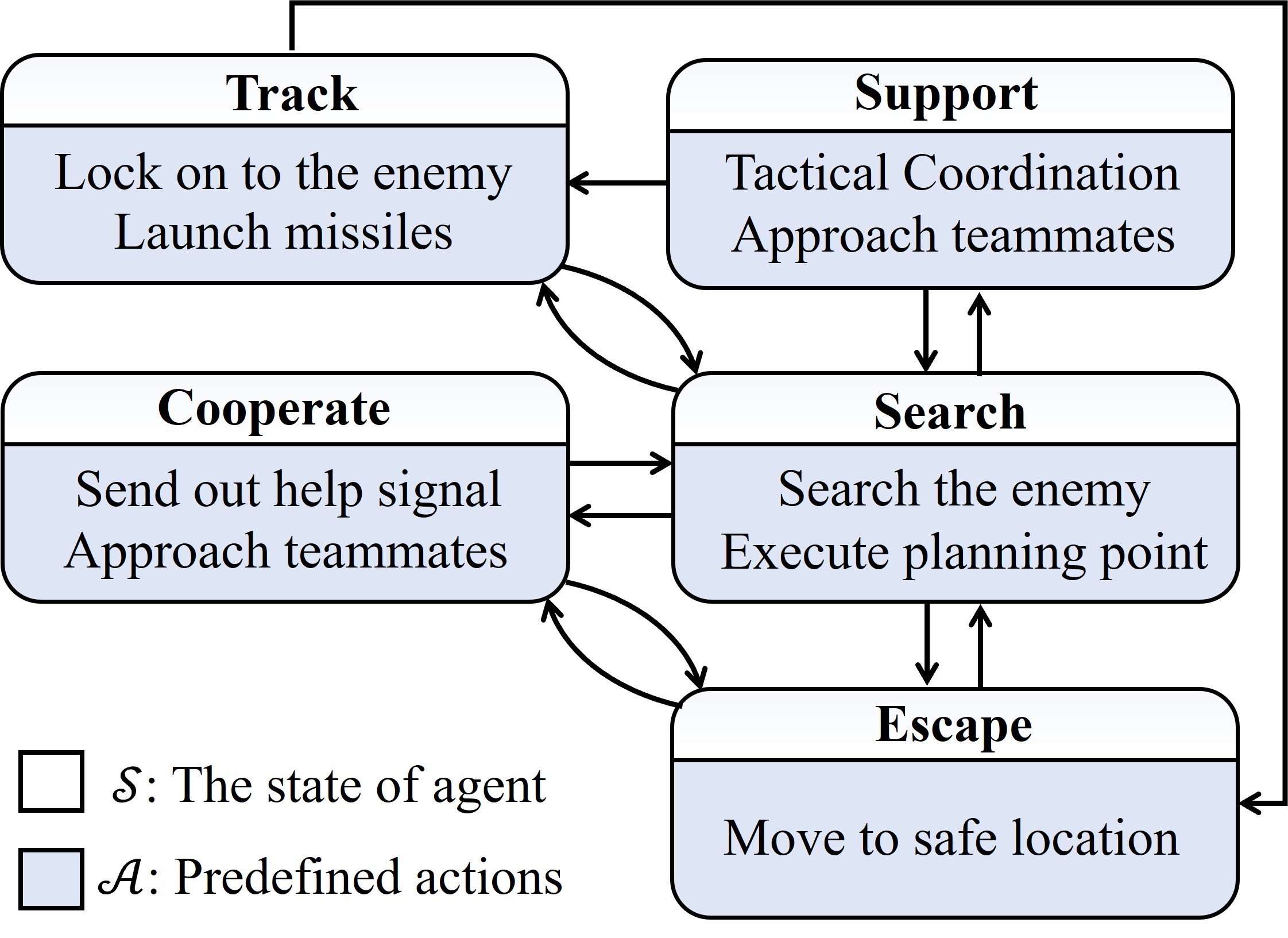}
    \caption{The agent state transition model includes five main states: \textit{SupportState}, \textit{SearchState}, \textit{TrackState}, \textit{EscapeState}, and \textit{CooperateState}.
    The arrows in the diagram represent the transition relationships between states. For instance, an arrow from state \(i\) to state \(j\) indicates the transition from state \(i\) to state \(j\).
    All state transitions strictly follow these rules.}
    \label{state-trans}
\end{figure}

FSM are utilized in agent decision--making for their ability to effectively represent discrete behaviors and transitions. 
To address uncertainty in dynamic environments, this framework is extended into a probabilistic finite state machine by incorporating a probabilistic transition matrix \( \boldsymbol{P} \).
The PFSM model is defined as a quintuple \( G = (\mathcal{S}, \mathcal{I}, \mathcal{A}, \boldsymbol{P}, \Lambda) \), where:
\begin{itemize}
\item \( \mathcal{S} \) is the finite set of states, which encompasses all possible states of the system. It is associated with a fixed transition model, as shown in Fig.\ref{state-trans}.
\item \( \mathcal{I} \) is the initial probability distribution function, which assigns an initial probability to each state. It is defined as \( \mathcal{I}: \mathcal{S} \rightarrow [0,1] \), and the total sum of probabilities equals 1, as expressed by:
\begin{equation}
    \sum\nolimits_{s \in \mathcal{S}} \mathcal{I}(s) = 1.
\end{equation}
\item \( \mathcal{A} \) is the mapping of each state to a set of possible actions, which indicates the actions available at state \( s \). The specific actions are detailed in Table \ref{definition of confrontation actions}.
\item \( \boldsymbol{P} \) is the transition probability matrix, which defines the probability of transitioning from state \( s \) to state \( s' \) when action \( a \) is taken. It is defined as \( \boldsymbol{P}: \mathcal{S} \times \mathcal{A}(s) \times \mathcal{S} \rightarrow [0,1] \), and the matrix \( \boldsymbol{P} \) satisfies the normalization condition:
\begin{equation}
    \sum\nolimits_{s' \in \mathcal{S}} \boldsymbol{P}(s, a, s') = 1, \quad \forall s \in \mathcal{S}, \quad \forall a \in \mathcal{A}.
\end{equation}
\item \( \Lambda \) is the set of parameters for the deep learning model, which parameterizes the transition probabilities.
\end{itemize}

\begin{table}[t]
\centering
\caption{Correspondence between States and Actions}
\label{definition of confrontation actions}
\begin{tabular}{>{\centering\arraybackslash}m{2.5cm}  >{\centering\arraybackslash}m{5cm}}
\toprule
State \( \mathcal{S} \) & Action \(\mathcal{A} \) \\
\midrule
\multirow{2}{*}{\textit{SearchState}} & Search the Enemy \\
& Execute Planning Point\\
\multirow{2}{*}{\textit{TrackState}} & Lock on to the Enemy \\
 & Launch missiles. \\
\textit{EscapeState}  & Move to Safe Location \\
\multirow{2}{*}{\textit{CooperateState}} & Send out Help Signal \\
 & Approach Teammates \\
\multirow{2}{*}{\textit{SupportState}}  & Tactical Coordination \\
 & Approach Teammates \\
\bottomrule
\end{tabular}
\end{table}

The state transition process in the PFSM model is essential for modeling the dynamic behaviors of agents in adversarial environments. In this probabilistic framework, the state transitions are governed solely by the transition probability matrix \( \boldsymbol{P} \).
When an agent is in state \( s_t \) and takes action \( a_t \), the probability of transitioning to state \( s_{t+1} \) is given by \( \boldsymbol{P}(s_t, a_t, s_{t+1}; \Lambda) \). The next state \( s_{t+1} \) is determined by
\begin{equation}
  s_{t+1} = \mathop{\arg\max}_{s' \in \mathcal{S}} \boldsymbol{P}(s_t, a_t, s'; \Lambda).
\end{equation}
This approach introduces a probabilistic component to the state transitions, ensuring that each transition becomes unique by incorporating the transition probability matrix \( \boldsymbol{P} \). 

\subsection{Compound Convolutional Network}

\begin{figure*}
    \centering
    \includegraphics[width=1.0\textwidth,height=0.2667447972157\textwidth]{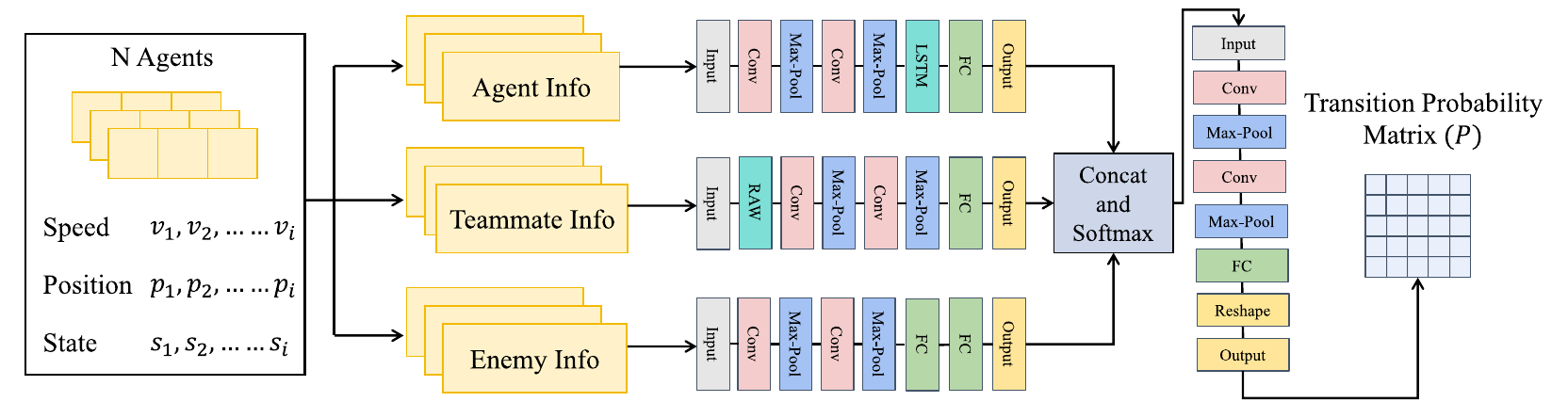}
    \caption{The network is designed to calculate the transition probability matrix \( \boldsymbol{P} \) for multi-agent systems by processing information from agents, teammates, and enemies. Agent information includes speed, position, and state data, which are input into convolutional layers followed by max-pooling, fully connected (FC) layers, and an LSTM layer to capture sequential dependencies. Outputs from these three information streams are concatenated and normalized using a softmax layer. The combined features are then reshaped and further processed through additional convolutional and FC layers to generate the final transition probability matrix \( \boldsymbol{P} \), which defines the probabilistic state transitions for the agents.}
    \label{net}
\end{figure*}

In the swarm confrontation scenarios, conflicts often arise due to incomplete information among agents, teammates, and enemies.
The proposed network addresses this issue by integrating diverse information streams to generate a transition probability matrix \( \boldsymbol{P} \).
By leveraging a multi--channel convolutional neural network, the architecture processes three types of information streams: agent-specific data, teammate data, and enemy data. 
Each stream independently extracts features through a dedicated subnetwork, ensuring modularity and scalability. 
The extracted features are then concatenated and further transformed to generate \( \boldsymbol{P} \), which serves as the foundation for decision-making and strategy optimization in the swarm.

The network takes input data that includes velocity, position, state, and time information for each agent. 
The velocity and position are represented as two-dimensional vectors, while the state is a discrete value that indicates current behavior of the agent. 
The network normalizes the velocity and position components using the mean and standard deviation calculated across all agents to maintain numerical stability. The state is converted into a one-hot vector to encode categorical information. Time steps are arranged sequentially to provide input suitable for temporal modeling.
The preprocessed data is organized into three separate streams: agent-specific data, teammate data, and enemy data. Each stream is processed independently by its corresponding network module to extract time--dependent features.
Each subnetwork begins with convolutional layers, which capture spatial dependencies and local patterns in the input data. These layers are responsible for extracting essential features related to velocity, position, and state, allowing the network to represent the spatial and temporal relationships inherent in each stream. The feature maps generated by the convolutional layers are then passed through max-pooling layers, which downsample the spatial dimensions of the data. This step retains the most salient information while reducing the computational complexity of subsequent operations. 
The reduced feature maps are processed by fully connected layers, which condense the high-dimensional representations into compact feature vectors. 
The extracted feature vector at each time step \( t \) can be expressed as follows:
\begin{equation}
\begin{cases}
\boldsymbol{h}_c^0 &= 0, \\
\boldsymbol{e}^t &= E_1(\boldsymbol{s}_p^t, W_1), \\
\boldsymbol{o}_c^t &= F(\boldsymbol{e}^t, W_2), \\
\boldsymbol{r}^t &= E_2(\boldsymbol{o}_c^t, W_3),
\end{cases}
\end{equation}
where \(E_1\), \(E_2\), and \(F\) represent the embedding networks at different stages of the subnetwork. 
\(\boldsymbol{h}_c^0 = 0\) initializes the feature map. 
\(\boldsymbol{e}^t\) is the encoded feature vector extracted from the input \(\boldsymbol{s}_p^t\) by \(E_1\) with parameters \( W_1 \). 
The intermediate vector \(\boldsymbol{o}_c^t\) is obtained by applying transformations to \(\boldsymbol{e}^t\) using \(F\) with \( W_2 \). 
\(\boldsymbol{r}^t\) is the compact output feature generated by \(E_2\) with \( W_3 \).

The AgentNet and EnemyNet modules incorporate LSTM layers to model temporal dependencies and allow the network to retain historical information about agent trajectories and observed enemy movements. The system encodes the input data at each time step, including velocity, position, and state, into high-dimensional feature representations through an embedding function. It then passes these representations to the LSTM layer, which updates its hidden states by integrating the encoded features with temporal information from previous time steps. The system transforms the updated hidden states into compact feature vectors through a fully connected layer. These vectors capture both spatial characteristics and temporal dynamics, enabling the network to represent the evolving behaviors of agents and enemies. The inclusion of LSTM layers enhances the network’s ability to account for historical patterns, which improves decision-making in dynamic swarm environments.

The TeammateNet module enhances inter--agent collaboration by dynamically weighting information shared among teammates. It employs a relational attention weighting mechanism to adapt feature representations based on the relevance of teammates’ information. This mechanism ensures that critical relationships among teammates are emphasized, while less relevant interactions are attenuated, and thereby improves the network's ability to model collaborative behaviors. At each time step \(t\), the system encodes the input data from teammates, including velocity, position, and state, into high-dimensional feature representations using the embedding function \(E_1\). It then computes the relational attention weight \(\alpha_{i,j}^t\) for each teammate \(j\) relative to the focal agent \(i\), based on their relative positions and states
\begin{equation}
\alpha_{i,j}^t = \frac{e^{-d(i, j)/\tau}}{\sum_{k} e^{-d(i, k)/\tau}},
\end{equation}
where \(d(i, j)\) quantifies the similarity or distance between agents \(i\) and \(j\), and \(\tau\) is a temperature parameter that controls the sensitivity of the attention weights. The attention weights are then used to aggregate teammate features, producing a weighted feature representation
\begin{equation}
\boldsymbol{e}_i^t = \sum_{j} \alpha_{i,j}^t \cdot E_1(\boldsymbol{s}_j^t, W_1),
\end{equation}
where \(\boldsymbol{s}_j^t\) represents the input data of teammate \(j\) at time step \(t\).
The aggregated feature representation \(\boldsymbol{e}_i^t\) is further processed through nonlinear transformations and fully connected layers to generate the final compact feature vector \(\boldsymbol{r}_i^t\). This representation captures both spatial dependencies and dynamic interactions among teammates, enabling the network to effectively model collaborative behaviors within the swarm.

The extracted feature vectors from AgentNet, TeammateNet, and EnemyNet are concatenated to form a unified representation for each agent. 
This concatenated feature vector captures the comprehensive spatial, temporal, and relational information required for decision-making. 
The concatenation operation is expressed as:
\begin{equation}
\boldsymbol{z} = \text{concat}\left( \boldsymbol{r}_\text{a}, \boldsymbol{r}_\text{t}, \boldsymbol{r}_\text{e} \right),
\end{equation}
where \( \boldsymbol{r}_\text{a} \), \( \boldsymbol{r}_\text{t} \), and \( \boldsymbol{r}_\text{e} \) represent the final outputs of the respective subnetworks. 
This unified representation is passed through the policy network, which generates parameters for the PFSM state transition matrix \( \boldsymbol{P} \). The matrix dynamically adapts based on the current feature representations, ensuring that decision-making reflects both cooperative and adversarial dynamics
\begin{equation}
\boldsymbol{P}(s_t, a_t, s_{t+1}) = f_{\Lambda}\left( \boldsymbol{z} \right),
\end{equation}
where \( f_{\Lambda} \) represents a transformation parameterized by the policy network's learnable weights. 
The integration of these components enables agents to adaptively update their state transitions based on contextual information, improving decision-making efficiency. By combining the strengths of AgentNet, TeammateNet, and EnemyNet, the framework provides a comprehensive understanding of agent states, teammate interactions, and opponent strategies, facilitating coordinated and adaptive behavior within the swarm.

\subsection{Reinforcement Learning}
To enhance the adaptability of agents in dynamic swarm adversarial environments, we integrate a reinforcement learning module into our decision-making framework. The reinforcement learning module employs the Proximal Policy Optimization (PPO) algorithm, leveraging the Actor-Critic architecture to optimize policies while ensuring stability and efficiency in learning.
The Actor network uses the PolicyNet, which combines the outputs from AgentNet, TeammateNet, and EnemyNet, to generate transition probabilities \( \pi_\theta(s_{t+1} \mid s_t) \) for the current state \( s_t \). By integrating these subnetworks, the agent captures comprehensive spatiotemporal and relational features, enabling more informed policy updates. 
Meanwhile, the Critic network evaluates the value function \( V^\pi(s_t, \boldsymbol{P}) \), incorporating the dynamic state transition probability matrix \( \boldsymbol{P} \), providing feedback to the Actor for strategy refinement.

At each time step \( t \), the agent observes external information and determines its next state \( s_{t+1} \in \mathcal{S} \) based on its current state \( s_t \) and the policy \( \pi_\theta \). The policy \( \pi_\theta \) incorporates the dynamic state transition probabilities \( \boldsymbol{P} \), enabling the agent to adapt to changes in the environment. The agent then receives a reward \( r_t \) associated with this state transition. The objective is to maximize the expected cumulative reward from the current state \( s_t \), defined as:
\begin{equation}
V^\pi(s_t, \boldsymbol{P}) = \mathbb{E}_{\pi} \left[ \sum\nolimits^{\infty}_{k=0} \gamma^k r_{t+k} \,\middle|\, s_t, \boldsymbol{P} \right],
\end{equation}
where \( \gamma \in [0,1] \) is the discount factor, balancing immediate and future rewards. 
The inclusion of \( \boldsymbol{P} \) enables the Critic network to eval uate state transitions dynamically, considering the probabilistic nature of the environment and the agent's strategic adjustments. 
PPO optimizes the policy by maximizing a clipped surrogate objective function
\begin{equation}
\begin{aligned}
L^{\text{CLIP}}(\theta) = \mathbb{E}_t \bigg[
&\min \Big(
\rho_t(\theta) \hat{A}_t, 
\text{clip}\big(\rho_t(\theta), 1 - \epsilon, 1 + \epsilon\big) \hat{A}_t
\Big) \\
&+ \lambda_1 \|\boldsymbol{P}\|_1 + \lambda_2 \|\boldsymbol{P} - \boldsymbol{P}^\text{t}\|_F^2
\bigg],
\end{aligned}
\end{equation}
where the importance sampling ratio \( \rho_t(\theta) \), the advantage function \( \hat{A}_t \), and
\begin{equation}
\rho_t(\theta) = \frac{\pi_\theta(s_{t+1} \mid s_t)}{\pi_{\theta_{\text{old}}}(s_{t+1} \mid s_t)},
\end{equation}
\begin{equation}
\hat{A}_t = r_t + \gamma V^\pi(s_{t+1}, \boldsymbol{P}) - V^\pi(s_t, \boldsymbol{P}).
\end{equation}
Here, \( \rho_t(\theta) \) represents the importance sampling ratio, measuring the relative probability of transitioning to state \( s_{t+1} \) under the current and previous policies. The advantage function \( \hat{A}_t \) quantifies the relative benefit of transitioning from \( s_t \) to \( s_{t+1} \), where a positive value indicates a favorable transition. The regularization terms enforce sparsity (\( \|\boldsymbol{P}\|_1 \)) and structural consistency (\( \|\boldsymbol{P} - \boldsymbol{P}^\text{t}\|_F^2 \)) in the state transition probability matrix \( \boldsymbol{P} \), controlled by the hyperparameters \( \lambda_1 \) and \( \lambda_2 \).

The Actor network parameters \( \varphi \) are updated to maximize the PPO objective, while the Critic network parameters \( w \) are updated to minimize the temporal difference (TD) error. The updates are given by
\begin{align}
\varphi_{t+1} &= \varphi_t + \eta_\varphi \nabla_\varphi L^{\text{CLIP}}(\varphi_t), \\
w_{t+1} &= w_t - \eta_w \nabla_w \left[ \frac{1}{2} \left( V_w(s_t, \boldsymbol{P}) - V_t \right)^2 \right],
\end{align}
where the target value \( V_t \) incorporates a correction term to account for state transition uncertainty
\begin{equation}
V_t = r_t + \gamma V_w(s_{t+1}, \boldsymbol{P}) + \eta \Delta \boldsymbol{P}(s_t, s_{t+1}),
\end{equation}
where \( \Delta \boldsymbol{P}(s_t, s_{t+1}) \) quantifies the variance of state transitions within the matrix \( \boldsymbol{P} \), and \( \eta \) is a scaling factor that determines the influence of this correction term.

By integrating the transition matrix \( \boldsymbol{P} \) into the PPO framework, the Actor and Critic networks dynamically adapt to the probabilistic nature of state transitions in swarm adversarial environments. This integration enhances robustness, ensuring informed decision-making that balances cooperation and confrontation in complex scenarios.

To address the challenges of deadlock and jitter in the PFSM--based decision-making framework, we propose a reward function that integrates multiple objectives to ensure efficient and stable agent behavior. This reward function combines task-related incentives with penalties designed to discourage undesirable behaviors, thus promoting consistent progress toward mission objectives. The overall reward at time step \( t \) is defined as:
\begin{equation}
\begin{split}
R(s_t, a_t, s_{t+1}) = R_{\text{t}}(s_t, a_t, s_{t+1}) - \lambda_{\text{d}} R_{\text{d}}(s_{t+1}) \\
- \lambda_{\text{j}} R_{\text{j}}(s_t, s_{t+1}),
\end{split}
\end{equation}
where \( R_{\text{t}}(s_t, a_t, s_{t+1}) \) represents the task-related reward, \( R_{\text{d}}(s_{t+1}) \) penalizes deadlock states, and \( R_{\text{j}}(s_t, s_{t+1}) \) addresses jitter-induced inefficiencies. The coefficients \( \lambda_{\text{d}} \) and \( \lambda_{\text{j}} \) balance the relative contributions of each component, ensuring that the agent remains focused on its objectives while avoiding undesirable behaviors.

The task-related reward \( R_{\text{t}}(s_t, a_t, s_{t+1}) \) incentivizes actions that align with the mission’s objectives. It encourages transitions that move the agent closer to its goal while penalizing inefficient or resource-intensive actions. The reward is given by:
\begin{equation}
R_{\text{t}}(s_t, a_t, s_{t+1}) = r_{\text{s}} \boldsymbol{P}(s_t, s_g) - r_{\text{c}}(a_t),
\end{equation}
where \( r_{\text{s}} > 0 \) reflects the reward for progressing toward the goal state \( s_g \), as measured by the transition probability \( \boldsymbol{P}(s_t, s_g) \). The term \( r_{\text{c}}(a_t) \) penalizes costly or suboptimal actions, ensuring resource-efficient behavior.

To prevent deadlock, where the agent becomes stuck in a state with no valid transitions, we introduce the penalty term \( R_{\text{d}}(s_{t+1}) \). Deadlock occurs when all outgoing transitions are zero except for a self-transition, indicating that the agent has no feasible paths forward. This penalty is defined as:
\begin{equation}
R_{\text{d}}(s_{t+1}) = r_{\text{d}}\mathbb{I}\left( \sum\nolimits_{j \neq i} \boldsymbol{P}(s_{t+1}, s_j) = 0 \right),
\end{equation}
where \( \mathbb{I}(\cdot) \) is an indicator function that activates when the agent is in a deadlock state, and \( r_{\text{d}} > 0 \) determines the penalty's magnitude. By penalizing such states, this term encourages the agent to avoid decisions leading to stagnation.

The reward function incorporates a penalty term \( R_{\text{j}}(s_t, s_{t+1}) \) to address jitter, which is characterized by frequent and inefficient oscillatory transitions between states. This penalty leverages the variance in the state transition probabilities \( \boldsymbol{P} \) to quantify instability. Specifically, jitter is penalized when the variance of the transition probabilities between states is high, indicating a lack of stability in decision-making. The penalty term is expressed as:
\begin{equation}
R_{\text{j}}(s_t, s_{t+1}) = r_{\text{j}}\textit{Var}\big(P(s_t, s_{t+1})\big) ,
\end{equation}
where \( \textit{Var}\big(P(s_t, s_{t+1})\big)\) represents the variance of the state transition probabilities \( P(s_t, s_{t+1}) \). 

By integrating these components, the reward function creates a balance between encouraging goal-directed behavior and discouraging inefficiencies. This approach results in a robust reward design that not only guides the agent toward its mission goals but also enforces stability and efficiency in its decision-making processes.

\section{Results and Analysis}
\label{sec:results_discussions}
This section presents simulations and real--world experiments to evaluate the proposed decision--making framework, highlighting its performance against baseline algorithms and analyzing the impact of each framework component.
\subsection{Experiment Settings}   
We design the swarm confrontation, and set the dynamic parameters of agents in the simulation as shown in Table \ref{Parameters}. 
The simulation environment and the application of the algorithm run on a computer equipped with an Intel(R) Xeon(R) W-2150B CPU, 64GB RAM, and an NVIDIA GeForce RTX 3090 GPU. 
The hyperparameters of the learning algorithm are shown in Table \ref{Hyperparameters}.
\begin{table}[t]
\centering
\caption{Simulation Environment Parameters}
\begin{tabular}{>{\centering\arraybackslash}m{1cm} >{\centering\arraybackslash}m{4.5cm} >{\centering\arraybackslash}m{1.5cm}}
\toprule
Feature & Description & Value \\
\midrule
\( r_{d} \) & The observation distance of agent & 2 m \\
\( r_{s} \) & The attack distance of agent & 1.5 m \\
\( \theta_d \) & The observation angle of agent & 360 degrees \\
\( \theta_s \) & The attack angle of agent & 80 degrees \\
\( m \) & Maximum number of missiles & 2 \\
\( \boldsymbol{v}_{\text{max}} \) & Maximum velocity of agent & 1.5 m/s \\
\( X \) & Width of the simulation area & 22 m  \\
\( Y \) & Height of the simulation area & 15 m \\
\bottomrule
\label{Parameters}
\end{tabular}
\end{table}
\begin{table}[t]
\centering
\caption{Learning Algorithm Hyperparameters}
\begin{tabular}{>{\centering\arraybackslash}m{3cm} >{\centering\arraybackslash}m{1.5cm}}
\toprule
Hyperparameter & Value \\
\midrule
Hidden Dimension & 512 \\
Actor Learning Rate & $1 \times 10^{-4}$ \\
Critic Learning Rate & $1 \times 10^{-3}$ \\
Lambda & 0.95 \\
Epsilon & 0.2 \\
Gamma & 0.98 \\
Number of Episodes & 100 \\
Maximum Steps & 512 \\
\bottomrule
\label{Hyperparameters}
\end{tabular}
\end{table}
\subsection{Agent Configuration}

The proposed decision-making framework is developed, trained, and evaluated within the context of the swarm adversarial environment to ensure its effectiveness.
The enemy agents use the finite state machine as the top control module. Each state corresponds to specific actions, which are the same under each state, as shown in Table \ref{definition of confrontation actions}.

The configuration of each agent is central to its performance in an adversarial environment. 
Each agent incorporates various sensors that provide crucial real--time data about the environment. 
They provide long-range detection of objects and obstacles and give the agent a highly reliable source of environmental data that is essential for dynamic decision--making and path planning.
The agent leverages the dynamic window approach (DWA) algorithm to compute the optimal linear and angular velocities, \(v\) and \(\omega\), for real-time navigation. The DWA algorithm evaluates candidate velocity pairs \( (v, \omega) \) within a dynamic window \( \mathcal{V}_d \), defined as:
\begin{equation}
\mathcal{V}_d = \left\{(v, \omega) \;\middle|\; 
\begin{aligned}
v &\in \left[ v_c - a_v^{\max} \Delta t, \, v_c + a_v^{\max} \Delta t \right], \\
\omega &\in \left[ \omega_c - a_\omega^{\max} \Delta t, \, \omega_c + a_\omega^{\max} \Delta t \right]
\end{aligned}
\right\},
\end{equation}
where \( v_c \) and \( \omega_c \) are the agent's current linear and angular velocities, \( a_v^{\max} \) and \( a_\omega^{\max} \) are the maximum linear and angular accelerations, and \( \Delta t \) represents the time step. This formulation dynamically adjusts the range of candidate velocities based on the motion states and constraints of agent.
Each candidate velocity pair \( (v, \omega) \) within \( \mathcal{V}_d \) is evaluated using an objective function \( G(v, \omega) \), which optimizes the trade-off between goal-oriented movement, obstacle avoidance, and trajectory smoothness. The optimal velocity pair \( (v^\ast, \omega^\ast) \) is then determined by
\begin{equation}
(v^\ast, \omega^\ast) = \mathop{\arg\max}_{(v, \omega) \in \mathcal{V}_d} G(v, \omega).
\end{equation}
By iteratively applying the computed optimal velocities \( (v^\ast, \omega^\ast) \), the agent achieves efficient navigation toward its target while avoiding obstacles, ensuring smooth and adaptive behavior in dynamic environments.

\subsection{Ablation Experiment}

\begin{figure*}
    \includegraphics[width=1.0\textwidth,height=0.4984219754529\textwidth]{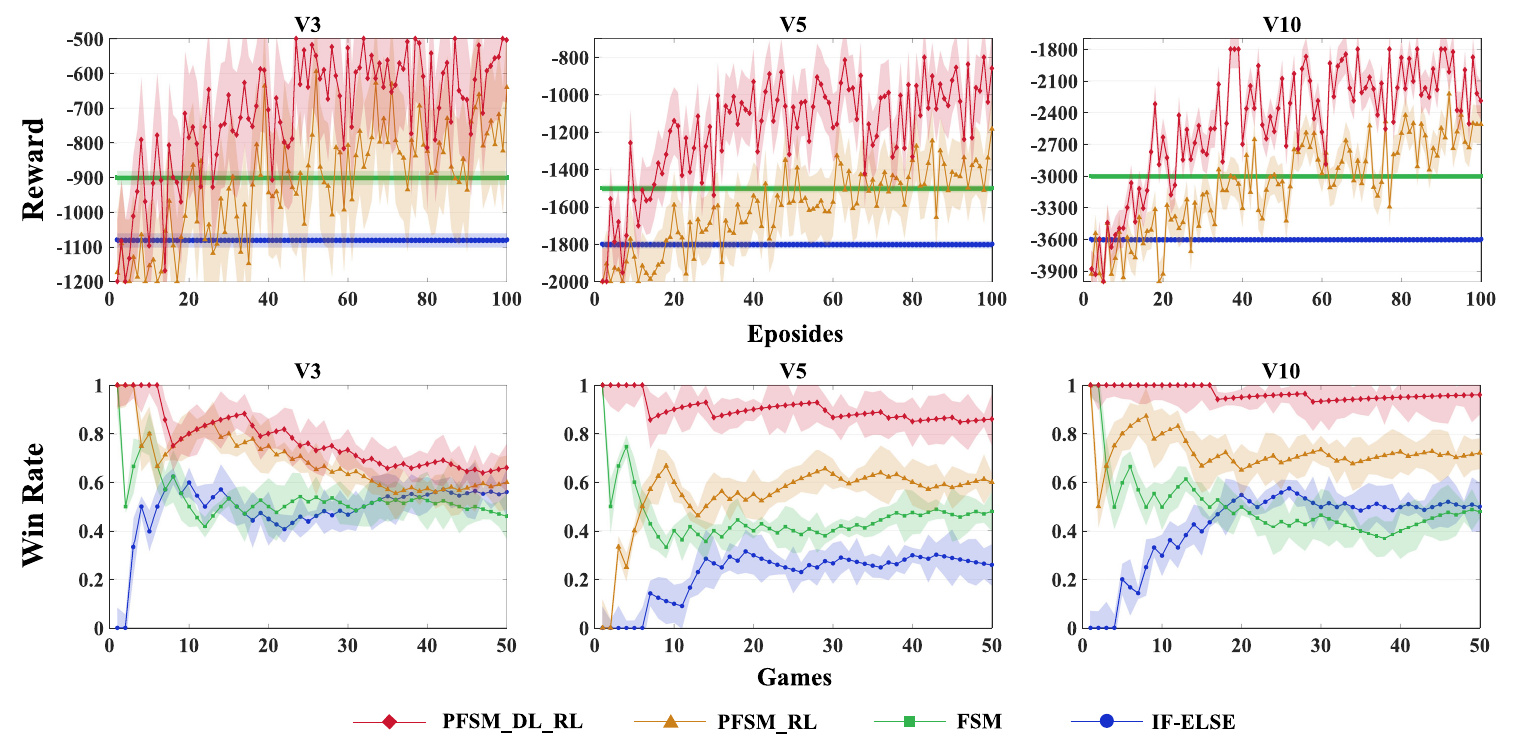}
    \caption{The ablation experiments evaluate four decision-making frameworks under different number of agents. The top subfigure shows the reward trends over 100 training episodes for agent configurations V3, V5, and V10. The bottom subfigure displays the win rate variations over 50 games for agent configurations V3, V5, and V10.  The shaded areas indicate the standard deviation of the results across repeated experiments.}
    \label{win_rate_plot}
\end{figure*}
This section analyzes the results of ablation experiments conducted under varying numbers of agents, focusing on the performance of four algorithms: the proposed algorithm, which integrates the PFSM with DRL; the PFSM combined with RL; the finite state machine; and a baseline based on if--else rules. The analysis evaluates both average rewards over 100 episodes and win rates over 50 games. The primary objective of the proposed algorithm is to address key limitations in traditional state machine-based approaches, including jitter and deadlock issues, while maintaining adaptability in dynamic, swarm adversarial environments.

The average reward trends, shown in the top row of the Fig. \ref{win_rate_plot}, highlight the superior learning efficiency and stability of the proposed PFSM--DRL algorithm. At lower agent counts, the proposed method achieves rapid convergence to high rewards, maintaining stable performance with minimal variance across episodes. This stability is particularly notable as it demonstrates the algorithm’s ability to mitigate the inherent instability of probabilistic state machines. In contrast, the PFSM--RL approach shows slower convergence and moderate rewards, reflecting its inability to effectively utilize state transitions without the representational power of DRL. The FSM algorithm exhibits even lower rewards, which fluctuate significantly due to frequent state oscillations and the lack of adaptability to dynamic adversarial environments. The if-else rule-based baseline performs the worst, with rewards stagnating at low levels across all episodes. Its rigid decision--making framework prevents it from learning or adapting, especially as the complexity of interactions increases. As the number of agents grows, the advantage of the proposed algorithm becomes more pronounced. While the baselines struggle to adapt, with FSM and if--else rules suffering from compounded deadlock and state oscillation issues, the proposed PFSM--DRL approach scales effectively, maintaining high rewards and stable performance even in the most complex multi-agent scenarios.

The win rate results, shown in the bottom row of the Fig. \ref{win_rate_plot}, further emphasize the robustness and scalability of the proposed method. Across all agent configurations, the PFSM--DRL algorithm consistently achieves near-perfect win rates, demonstrating its ability to handle adversarial dynamics efficiently. With smaller agent counts, the proposed method quickly converges to a win rate close to 1.0 and sustains this level throughout the evaluation. In comparison, the PFSM--RL approach achieves higher win rates than the FSM and if-else rules but suffers from significant instability. The lack of deep representational learning in this approach limits its capacity to adapt effectively to adversarial strategies, leading to frequent state transitions that fail to optimize decision-making. The FSM algorithm performs worse, with win rates heavily affected by state oscillations and deadlocks that prevent effective task execution. These issues are particularly exacerbated as the number of agents increases, where the rigidity of FSM transitions fails to account for the increased environmental complexity. The if--else rule baseline consistently achieves the lowest win rates, with no signs of improvement, as it lacks the capacity to adapt or recover from adversarial challenges.

Moreover, the scalability of the proposed algorithm is evident in its consistent performance as the number of agents increases. While FSM and traditional methods exhibit increased instability and performance degradation under higher agent counts, the PFSM--DRL algorithm maintains stable and superior performance. This robustness is attributed to its ability to generalize learned policies across different agent interactions and environmental complexities.

\begin{figure*}
    \includegraphics[width=1.0\textwidth,height=0.3079761683880\textwidth]{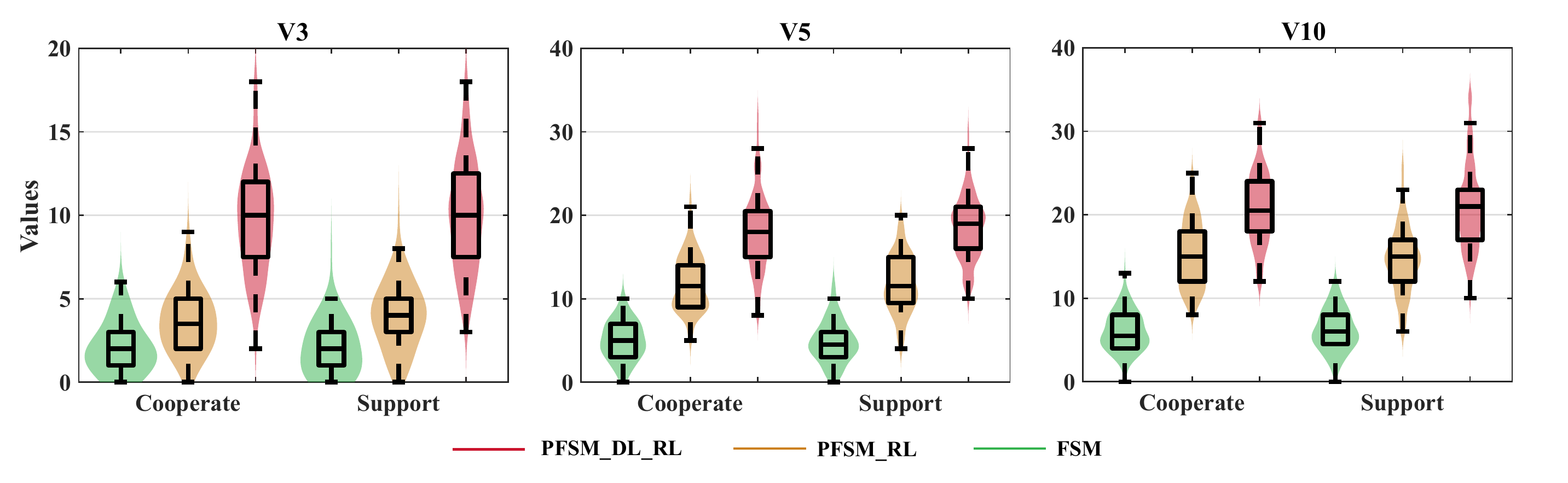}
    \caption{The average distribution of state values for \textit{Cooperate} and \textit{Support} states is presented across three agent configurations: V3, V5, and V10. The vertical spread represents the variability in state values, while the box plots within each violin indicate the interquartile range and median values.}
    \label{vio}
\end{figure*}

Building upon the earlier analysis of rewards and win rates, further insights into agent behaviors are revealed through Fig. \ref{vio}. 
The distributions illustrate how effectively each method fosters cooperative and supportive behaviors across configurations with varying numbers of agents.

In the cooperate and support state, the proposed PFSM--DRL algorithm exhibits significantly higher and wider distributions compared to the PFSM--RL and FSM baselines. This indicates that agents employing the proposed method not only engage in cooperative behavior more frequently but also sustain these interactions with greater consistency and intensity, particularly as the agent count increases.  
The PFSM--DRL method achieves a balanced overlap between cooperate and support distributions, reflecting its ability to dynamically switch between these states based on environmental demands. In comparison, the FSM and PFSM--RL baselines fail to achieve this balance, as seen in their more disjointed distributions and higher variance, which stem from limitations in adaptability and state transition efficiency.

\subsection{Real World Experiment} 
We evaluate the feasibility and effectiveness of the proposed framework by conducting experiments with real unmanned ground vehicles (UGVs) as test platforms. 
In these experiments, the UGVs are divided into two teams, referred to as the red and blue teams. 
The distinction between the teams lies in their external appearance: the red UGVs display a yellow base with red stripes, while the blue UGVs display a blue base with blue stripes. 
This visual differentiation facilitates tracking and identification during the experiments.
To eliminate bias caused by hardware differences, both teams use identical platform configurations, including the same sensors, actuators, and control modules. 
\begin{figure}
    \centering   
    \includegraphics[width=0.5\textwidth,height=0.1449632196486\textwidth]{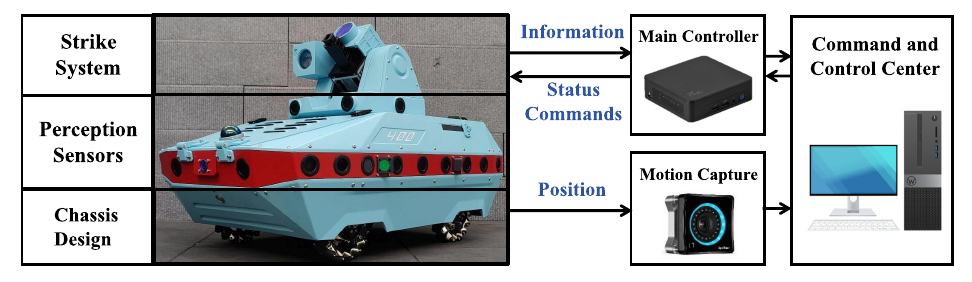}
    \caption{UGV system has three main subsystems: the Strike System (2D PTZ and Laser Strike), Perception Sensors (surround-view camera and LiDAR), and Chassis Design (Mecanum wheels for omnidirectional movement). The UGV’s Main Controller exchanges status and information with the Command and Control Center, while a Motion Capture system provides precise position data. This setup enables coordinated control and effective operation in complex environments.}
    \label{real-introduction}
\end{figure}

Each UGV, as shown in Fig. \ref{real-introduction}, comprises three main components. 
The topmost section is the striking system, which integrates a laser emitter mounted on a two-dimensional gimbal for targeted operations. The middle section houses the perception sensors, including a panoramic camera and radar, enabling a comprehensive reconnaissance capability. The chassis references classic vehicle frame designs, combining longitudinal and transverse beams, and is equipped with Mecanum wheels to enhance maneuverability and adaptability in complex environments.
The main control system of UGV is powered by the NUC13ANKI5, featuring a 13th generation Core i5-1340P CPU and running Ubuntu 20.04 with ROS Noetic. This configuration meets the computational demands of real--time decision--making and control tasks. The probabilistic finite state machine, embedded in the main control system, governs state transitions and actions based on expert-defined rules. The system communicates observational data with the command and control center, which processes this information using the proposed decision-making framework. The center sends back optimized state transition strategies to ensure coordinated and efficient team operations. 
The red team uses the proposed decision-making framework, which selects states based on observational data through a probabilistic finite state machine. 
In contrast, the blue team relies on a predefined rule-based strategy implemented with a finite state machine, where state transitions follow fixed rules.
Table \ref{definition of confrontation actions} outlines the actions corresponding to each state, ensuring transparency in the control logic.

We organize the real--world experiment into two sections, corresponding to random initial positioning ranges for the red and blue teams and place obstacles in the center to create a physical boundary, which prevents premature interactions when random positioning locates the UGVs too close to each other. 
A motion capture system tracks the precise positions of all UGVs in real time and transmits location data to the command and control center, ensuring seamless integration of spatial information into decision-making processes.

\begin{figure*}
    \includegraphics[width=1\textwidth,height=0.3773903856312\textwidth]{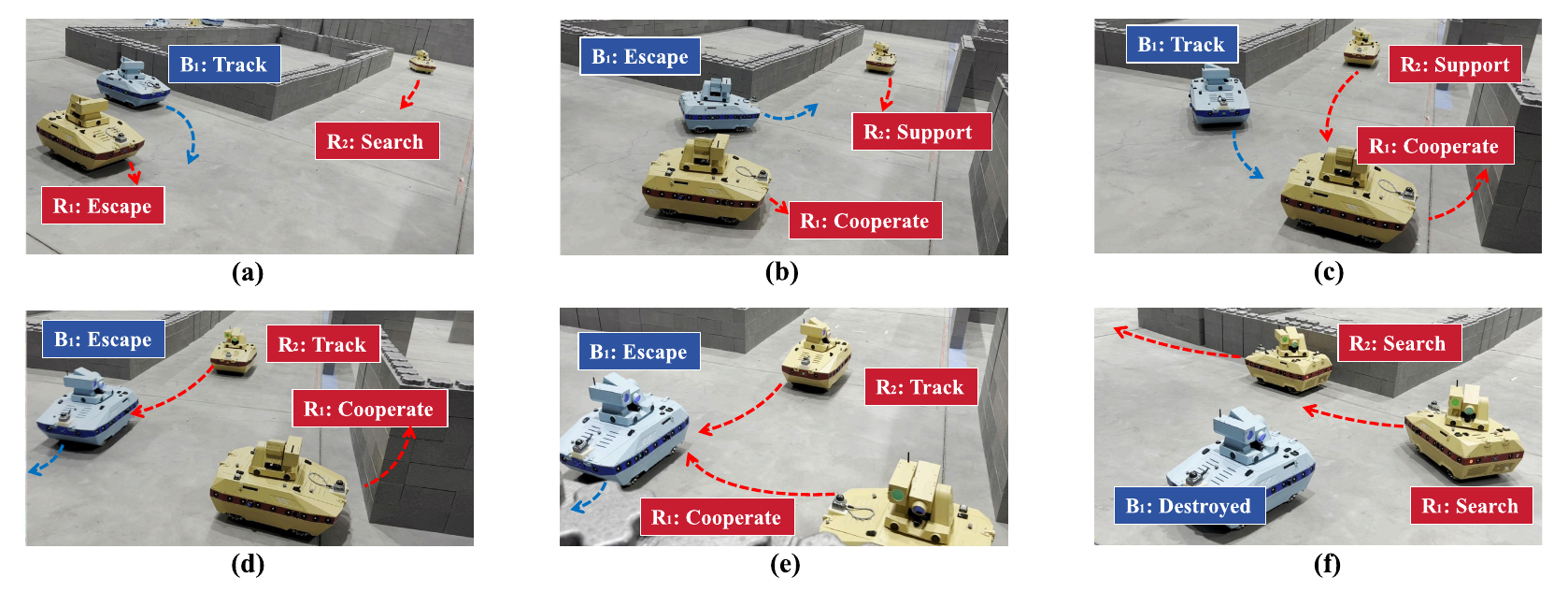}
    \caption{Dynamic state transitions and interactions in the swarm adversarial environment. 
    The scenarios depict coordination and competition between the blue team ({$B_1$}) and the red team ({$R_1$}, {$R_2$}). 
    In (a)–(c), the red UGVs counter the tracking state of blue UGV by employing cooperative and supportive strategies. 
    In (d)–(f), the blue UGV transitions to escape, prompting the red agents to intensify their pursuit, which results in the elimination of the blue UGV. 
    Dashed arrows represent movement trajectories, while state labels indicate the strategic decisions made by each agent.}
    \label{real-world}
\end{figure*}
\begin{figure*}
    \includegraphics[width=1\textwidth,height=0.3506692843119\textwidth]{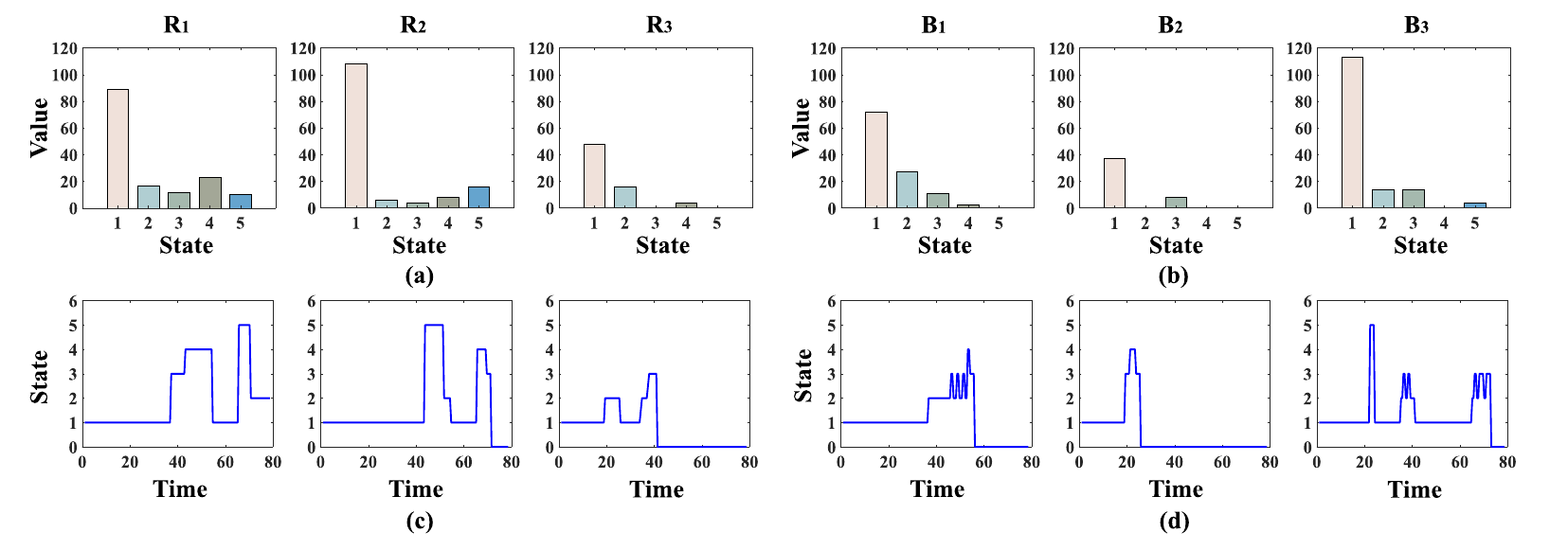 }
    \caption{State distributions and temporal transitions of agents in a swarm adversarial environment. Part (a) and (b) present the overall state distributions of agents across five predefined states: \textit{Search} (1), \textit{Track} (2), \textit{Escape} (3), \textit{Cooperate} (4), and \textit{Support} (5). Part (c) and (d) show the temporal evolution of agent states, revealing frequent transitions between states over time. These transitions highlight the adaptability and responsiveness of agents to dynamic adversarial conditions. }
    \label{real-world-exp}
\end{figure*}

We deploys six UGVs in a real--world experiment, assigning three to each team. The experiment captures six keyframes during one adversarial confrontation to compare the two decision--making methods, as shown in Fig.~\ref{real-world}. Fig~\ref{real-world-exp} provides statistical data for this experiment, offering insights into the behaviors and interactions of the UGVs throughout each phase.

The adversarial encounter begins with \(B_1\) detecting \(R_1\) during the initial confrontation phase. Based on the combat advantage, \(B_1\) transitions to the track state, while \(R_1\) switches to the escape state to evade tracking. \(R_1\) moves into a cooperative state, prompting \(R_2\), the nearest red UGV, to transition from its initial search state to the support state. This coordination enables \(R_1\) and \(R_2\) to align their actions effectively, improving their strategic positioning.

During the middle phase, \(R_1\) and \(R_2\) simultaneously enter the observation range of \(B_1\). Rapid changes in combat advantage cause \(B_1\) to oscillate between the track and escape states, which disrupts stable decision-making. These repeated state transitions, exacerbated by dynamic triggers, prevent \(B_1\) from stabilizing its strategy. Without support from other blue UGVs, \(B_1\) eventually loses its combat capability and is destroyed. Figure~\ref{real-world} demonstrates the progression of state transitions across the confrontation, cooperation, and final phases.

To further analyze state dynamics, Figs.~\ref{real-world-exp}(a)--(b) present histograms summarizing the occurrence of each state during the experiment. The results demonstrate that \( R_1 \) and \( R_2 \) spend the majority of time in cooperative and support states, which highlights their ability to synchronize actions effectively. In contrast, \( B_1 \) spends minimal time in cooperate or support states due to unstable decision-making under pressure, emphasizing its vulnerability when operating without team coordination.
The timeline of the adversarial process, as shown in Figs.~\ref{real-world-exp}(c)--(d), reveal additional insights. Between \( 35 \, \text{s} \) and \( 55 \, \text{s} \), \( R_1 \) and \( R_2 \) demonstrate smooth state transitions and maintain coordinated behavior. At \( 42 \, \text{s} \) and \( 55 \, \text{s} \), \( B_1 \) jitters between track and escape states, eventually resulting in its destruction by \( R_2 \).
Overall, the experiment highlights the dynamic decision-making capabilities of our framework. The ability to balance confrontation and cooperation ensures effective coordination among red UGVs, while blue UGVs struggle with state stability under multiple competing triggers. The accompanying video provides a detailed visualization of the entire process, further validating the real-world performance and robustness of our method.

\section{Conclusion}
\label{sec:conclusion} %
This paper has addressed decision jitter and deadlock in swarm confrontation by proposing a decision-making framework. 
It has used probabilistic finite state machines to convert condition matrices into transition probability matrices, optimizing them with conditional features to eliminate jitter and deadlock. 
The framework has introduced a multi-objective compound convolutional network that prioritizes agent cooperation to improve win rates. 
By assessing state transition probabilities from deep learning networks, the PPO algorithm has enhanced decision-making with reward settings aligned with human preferences. 
Experimental results have shown that our method outperforms the baseline in rewards and win rates, achieving over ninety percent win rates in large swarms. 
Ablation studies have validated the effectiveness of each component and transferring models trained on small swarms to large ones has demonstrated the generalization capability. 
After extensive training, our method has enabled agents to make effective decisions in swarm confrontations, making it suitable for deployment in real--robot systems.


%

\appendices


\bibliographystyle{IEEEtran}
\bibliography{IEEEabrv,tvtref}

\ifCLASSOPTIONcaptionsoff
  \newpage
\fi

\end{document}